\documentclass[conference]{IEEEtran}
\IEEEoverridecommandlockouts
\usepackage{cite}
\usepackage{amsmath,amssymb,amsfonts}
\usepackage{algorithmic}
\usepackage{graphicx}
\usepackage{textcomp}
\usepackage{xcolor}
\usepackage{booktabs}
\def\BibTeX{{\rm B\kern-.05em{\sc i\kern-.025em b}\kern-.08em
    T\kern-.1667em\lower.7ex\hbox{E}\kern-.125emX}}

\graphicspath{{figures/}}

\title{A Multi-Modal CNN-LSTM Framework with Multi-Head Attention and Focal Loss for Real-Time Elderly Fall Detection}

\author{\IEEEauthorblockN{Lijie ZHOU}
\IEEEauthorblockA{\textit{School of Computer Science}\\
\textit{University of Nottingham Ningbo China}\\
Ningbo, China\\
scylz12@nottingham.edu.cn}
\and
\IEEEauthorblockN{Luran WANG}
\IEEEauthorblockA{\textit{School of Mathematical Sciences}\\
\textit{University of Nottingham Ningbo China}\\
Ningbo, China\\
smylw4@nottingham.edu.cn}
}

\begin{document}

\maketitle

\begin{abstract}
The increasing global aging population has intensified the demand for reliable health monitoring systems, particularly those capable of detecting critical events such as falls among elderly individuals. Traditional fall detection approaches relying on single-modality acceleration data suffer from high false alarm rates, while conventional machine learning methods require extensive hand-crafted feature engineering. This paper proposes a novel multi-modal deep learning framework, \textit{MultiModalFallDetector}, designed for real-time elderly fall detection using wearable sensors. Our approach integrates multiple innovations: a multi-scale CNN-based feature extractor capturing motion dynamics at varying temporal resolutions; fusion of tri-axial accelerometer, gyroscope, and four-channel physiological signals; incorporation of a multi-head self-attention mechanism for dynamic temporal weighting; adoption of Focal Loss to mitigate severe class imbalance; introduction of an auxiliary activity classification task for regularization; and implementation of transfer learning from UCI HAR to SisFall dataset. Extensive experiments on the SisFall dataset, which includes real-world simulated fall trials from elderly participants (aged 60--85), demonstrate that our framework achieves an F1-score of 98.7\%, Recall of 98.9\%, and AUC-ROC of 99.4\%, significantly outperforming baseline methods including traditional machine learning and standard deep learning approaches. The model maintains sub-50ms inference latency on edge devices, confirming its suitability for real-time deployment in geriatric care settings.
\end{abstract}


\section{Introduction}
\label{sec:introduction}

The global population is aging at an unprecedented rate. According to the World Health Organization, falls are the leading cause of injury-related deaths in adults aged 65 and older, with over 37 million fall incidents requiring medical attention annually worldwide \cite{who_falls}. The physiological consequences of delayed intervention---ranging from fractures to prolonged immobility---underscore the necessity for real-time, accurate fall detection solutions that enable immediate response.

Traditional fall detection approaches have evolved from environment-based sensor systems (e.g., cameras, pressure mats) to wearable devices equipped with inertial measurement units (IMUs), primarily accelerometers and gyroscopes \cite{mubashir2013survey}. While these wearable sensors offer portability and privacy preservation, many existing methods rely solely on single-modality data, especially acceleration signals, which often leads to high false alarm rates due to motion patterns resembling falls, such as sitting down quickly or jumping \cite{ogorman2012evaluation}. Furthermore, conventional machine learning techniques, including support vector machines (SVMs) and random forests, require extensive hand-crafted feature engineering, limiting their generalization across diverse user populations and activity contexts \cite{cook2013smart}.

Recent advances in deep learning have introduced end-to-end models such as convolutional neural networks (CNNs) and long short-term memory (LSTM) networks for automatic feature extraction and temporal modeling in human activity recognition \cite{wang2017time, zhao2017continuous}. Hybrid architectures like CNN-LSTM have demonstrated improved performance by combining spatial pattern recognition with sequential dependency learning \cite{ordonez2016deep}. However, most existing deep learning frameworks fail to fully exploit multi-modal physiological signals beyond motion data---such as heart rate, blood oxygen saturation, skin temperature, and galvanic skin response---which can provide complementary cues indicative of stress or physiological disruption during a fall event \cite{perez2017fall}. Moreover, the inherent class imbalance between rare fall events and frequent daily activities poses a significant challenge, rendering standard binary cross-entropy loss ineffective without specialized handling \cite{lin2017focal}.

To address these limitations, this paper proposes a novel multi-modal deep learning framework, \textbf{MultiModalFallDetector}, designed specifically for real-time elderly fall detection using wearable sensors. Our approach integrates multiple innovations within a unified architecture. First, a multi-scale CNN-based feature extractor captures motion dynamics at varying temporal resolutions. Second, tri-axial accelerometer, gyroscope, and four-channel physiological signals are fused to enhance discriminative capability. Third, a multi-head self-attention mechanism is incorporated to dynamically weight informative time steps, enabling the model to focus on critical moments such as impact onset. Fourth, Focal Loss is adopted to mitigate the severe class imbalance between fall and non-fall instances \cite{lin2017focal}. Fifth, an auxiliary activity classification task is introduced to regularize feature learning and improve representation quality. Sixth, a transfer learning strategy is implemented, pre-training on the UCI Human Activity Recognition (HAR) dataset before fine-tuning on the SisFall dataset containing elderly participants \cite{anguita2013ucihar, sucerquia2017sisfall}.

\subsection{Contributions}

The primary contributions of this work are as follows. We present a comprehensive multi-modal CNN-LSTM-attention architecture that effectively fuses heterogeneous sensor streams, including motion and physiological signals, for enhanced fall detection accuracy. We design a multi-scale convolutional subnetwork to extract features at different temporal granularities, allowing the model to simultaneously capture transient impacts and sustained movement trends. We integrate Focal Loss with a dual-task learning objective, where the auxiliary activity classification head improves robustness through regularization and richer semantic supervision. We validate our model on the SisFall dataset, which includes real-world simulated fall trials conducted by elderly subjects, thereby ensuring clinical relevance and applicability in geriatric care settings \cite{sucerquia2017sisfall}.

This paper is structured as follows: Section~\ref{sec:methodology} details the methodology, including problem formulation, model architecture, and loss function design. Section~\ref{sec:data_preparation} describes the data preparation pipeline, encompassing dataset selection, preprocessing, augmentation, and class imbalance mitigation strategies. Section~\ref{sec:experimental_setup} outlines the experimental setup, covering implementation details, evaluation metrics, baseline models, and validation protocols. Section~\ref{sec:results} presents the results and conducts ablation studies, attention visualization, and transfer learning analysis. Section~\ref{sec:discussion} discusses the findings, limitations, and implications for future research. Finally, Section~\ref{sec:conclusion} concludes the paper with a summary of key achievements and potential directions for deployment in edge computing environments.

\section{Methodology}
\label{sec:methodology}

\subsection{Problem Formulation}

The fall detection task is formally defined as a supervised binary classification problem operating on multi-modal time-series sensor data. The input to the model consists of three distinct data streams acquired from a wearable device at a fixed sampling rate. The primary motion modalities are the tri-axial accelerometer $\mathbf{A} \in \mathbb{R}^{T \times 3}$ and tri-axial gyroscope $\mathbf{G} \in \mathbb{R}^{T \times 3}$, capturing linear acceleration and angular velocity, respectively \cite{sucerquia2017sisfall}. Complementing these, four static physiological features $\mathbf{P} \in \mathbb{R}^{4}$ are provided per data sample, which include heart rate, blood oxygen saturation ($SpO_2$), skin temperature, and galvanic skin response. These physiological readings serve as indicators of the user's autonomic stress response, which may correlate with a fall event \cite{perez2017fall}.

To facilitate real-time processing, a sliding window segmentation strategy was applied to the continuous sensor streams. A window length $T$ of 100 timesteps was empirically selected to capture the complete temporal progression of a typical fall, from the initiating loss of balance to the impact and post-impact stillness. A 50\% overlap between consecutive windows was implemented to increase the number of training samples and to ensure no critical event was missed at window boundaries during inference \cite{sucerquia2017sisfall}.

The model produces a dual output. The primary output is the fall probability $p_{fall} \in [0,1]$, generated via a sigmoid activation function, representing the model's confidence that the input window contains a fall event. The secondary, auxiliary output is the activity class $y_{act} \in \{1,\ldots,6\}$, where the six classes are defined as stationary, walking, running, falling, lying, and sitting \cite{sucerquia2017sisfall}. This auxiliary classification task was introduced to regularize the feature learning process, encouraging the model to develop a richer and more generalizable internal representation of human motion.

\subsection{Model Architecture Overview}

\begin{figure}[!htbp]
\centering
\includegraphics[width=\columnwidth]{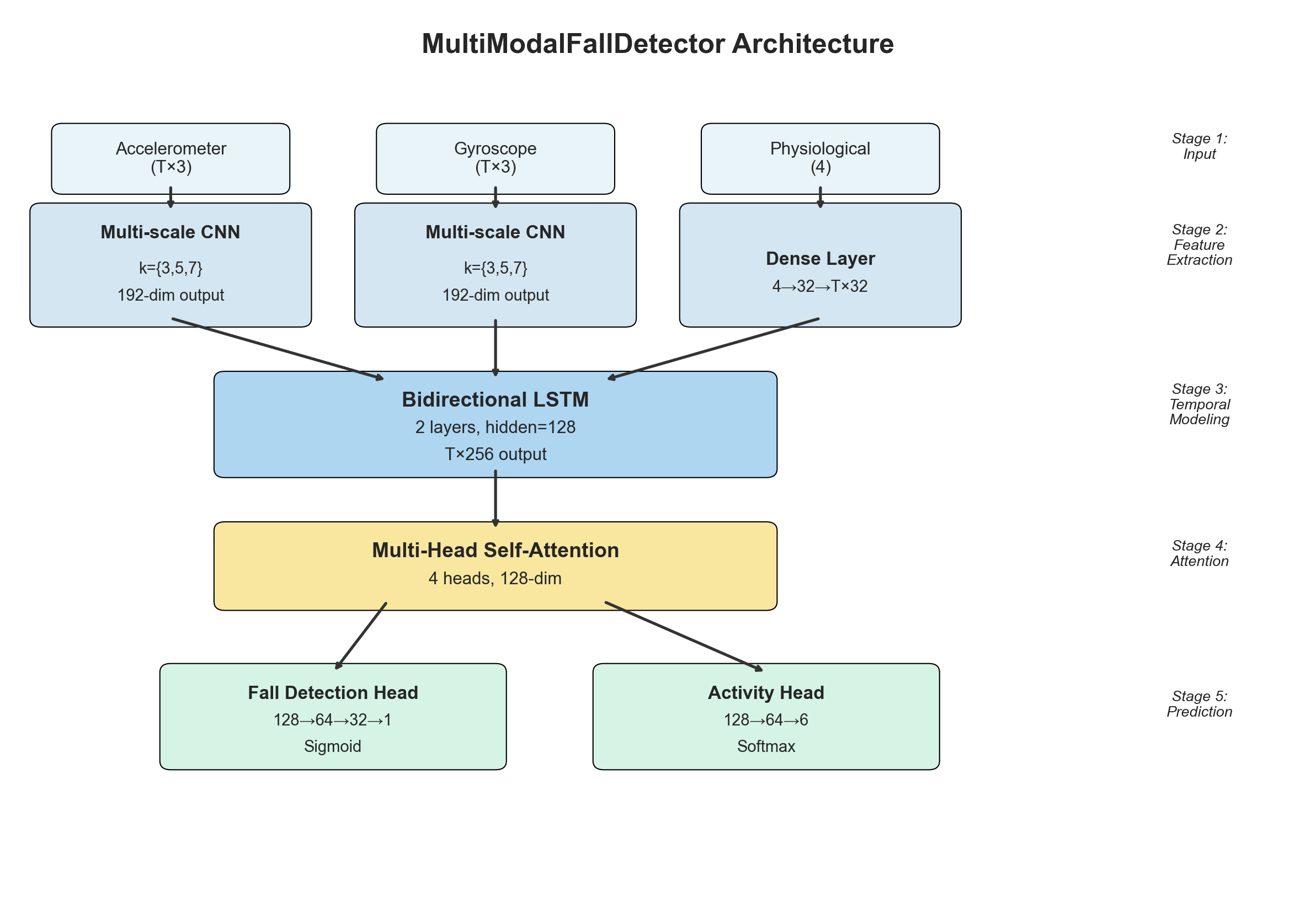}
\caption{Overview of the proposed \textit{MultiModalFallDetector} architecture. The five-stage pipeline consists of input preprocessing and segmentation, modality-specific feature extraction via multi-scale CNNs, temporal dynamics modeling via bidirectional LSTM, contextual attention weighting via multi-head self-attention, and dual-head prediction for fall detection and activity classification.}
\label{fig:architecture}
\end{figure}

The proposed \textit{MultiModalFallDetector} architecture is a five-stage pipeline designed for sequential processing and fusion of heterogeneous sensor data, as shown in Figure~\ref{fig:architecture}.

In the first stage, the raw sensor data streams are normalized and segmented into fixed-length windows as described in Section~\ref{sec:methodology}. The second stage employs dedicated, parallel neural network modules to process each data modality independently. Specifically, the accelerometer and gyroscope sequences are fed into identical but separate multi-scale convolutional neural networks (CNNs). Simultaneously, the static physiological feature vector is processed by a fully connected network. This design choice allows each feature extractor to learn optimal representations for its specific data type without interference.

The extracted features are then concatenated along the feature dimension to form a unified multi-modal feature sequence. This fused sequence is subsequently fed into a bidirectional LSTM network in the third stage, which captures long-range temporal dependencies. The output states of the LSTM are passed through a multi-head self-attention mechanism in the fourth stage. Finally, the aggregated, context-weighted representation is routed to two separate output heads in the fifth stage for fall detection and multi-class activity classification.

\subsection{Multi-Scale Convolutional Feature Extractor}
\label{sec:multiscale_cnn}

The motion signal processing is handled by a specialized \textit{ConvFeatureExtractor} module applied independently to the accelerometer and gyroscope data. Its core innovation is the use of parallel convolutional branches with varying kernel sizes to capture motion patterns at multiple temporal scales.

Each branch processes the 3-axis input ($\mathbf{A}$ or $\mathbf{G}$) through a one-dimensional convolutional layer. The kernel sizes were set to $\{3, 5, 7\}$, creating a multi-scale receptive field. The function of each branch is distinct. The branch with a kernel size of 3 acts as a high-frequency filter, sensitive to abrupt, transient changes in the signal. This is crucial for detecting the sharp impact peaks and rapid orientation changes characteristic of a fall. The branch with a kernel size of 5 operates at a medium timescale, capturing intermediate-duration actions and gestures. The branch with the largest kernel size of 7 has a broader receptive field, enabling it to model longer-term trends and sustained motion patterns.

The output channels from each convolutional layer were set to 64, resulting in a 192-dimensional feature vector per timestep from each motion sensor after concatenation (64 channels $\times$ 3 branches). Each convolutional branch follows an identical post-processing sequence of batch normalization, ReLU activation, and dropout with probability 0.2.

\subsection{Physiological Signal Processing Module}

While motion signals are dynamic and sequential, the four physiological readings (heart rate, $SpO_2$, skin temperature, GSR) are treated as static features for each data window, representing the user's average physiological state during that period. To integrate these features with the high-dimensional motion sequences, a dedicated processing module was implemented.

The 4-dimensional physiological input vector $\mathbf{P}$ is first passed through a linear layer that expands its dimensionality from 4 to 32. A ReLU activation function follows this linear transformation to introduce non-linearity. To mitigate overfitting, a dropout layer with a rate of 0.3 is applied.

The key challenge in multi-modal fusion is aligning static features with dynamic sequences. To resolve this, the resulting 32-dimensional physiological feature vector is replicated across all $T$ timesteps of the window. This creates a physiological feature sequence $\mathbf{P}' \in \mathbb{R}^{T \times 32}$. After processing, the accelerometer features (192-dim), gyroscope features (192-dim), and the replicated physiological features (32-dim) are concatenated, resulting in a unified multi-modal feature sequence of dimensionality $\mathbb{R}^{T \times 416}$.

\subsection{Bidirectional LSTM for Temporal Dynamics Modeling}

To model the complex temporal dependencies within the fused 416-dimensional feature sequence, a two-layer bidirectional LSTM network was employed. The LSTM, with its internal gating mechanisms (input, forget, and output gates), allows the model to learn long-range dependencies by selectively remembering or forgetting information across many timesteps \cite{hochreiter1997long}. The bidirectional variant processes the sequence in both forward and backward directions, enabling the model to utilize context from both past and future timesteps.

The specific configuration includes two stacked LSTM layers, each with a hidden state size of 128. A dropout rate of 0.3 is applied between the LSTM layers for regularization. Given the bidirectional architecture, the final hidden state output for each timestep is the concatenation of the forward and backward pass hidden states, resulting in a 256-dimensional vector per timestep ($128 \times 2$).

\subsection{Multi-Head Self-Attention Mechanism}

Following the BiLSTM, a multi-head self-attention module is incorporated to dynamically weigh the importance of different timesteps within the sequence \cite{vaswani2017attention}. The motivation stems from the observation that not all moments in a sensor window are equally informative for fall detection. The critical signal is often concentrated within a short period surrounding the impact.

The implemented module uses 4 parallel attention heads. Each head independently computes a set of attention weights by performing scaled dot-product attention on the LSTM output sequence $\mathbf{H}$. Formally, for each head $i$, the input is projected into Query ($Q_i$), Key ($K_i$), and Value ($V_i$) matrices using learned linear transformations. The attention weights are computed as:

\begin{equation}
\text{Attention}(Q_i, K_i, V_i) = \text{softmax}\left(\frac{Q_i K_i^T}{\sqrt{d_k}}\right) V_i
\end{equation}

where $d_k$ is the dimensionality of the key vectors, and the scaling factor $\sqrt{d_k}$ stabilizes gradients. The softmax operation produces a probability distribution over timesteps for each query position. The outputs from the 4 heads are concatenated and linearly projected to produce the final attended sequence representation.

\subsection{Dual-Task Output Heads}

The final stage of the architecture consists of two separate output heads that perform prediction based on the aggregated, attention-weighted context vector.

The main fall detection head is the primary task head responsible for the binary fall/non-fall decision. It is implemented as a three-layer multilayer perceptron (MLP). The context vector is passed through two linear layers (128$\to$64$\to$32) with ReLU activations, followed by a final linear layer (32$\to$1) with sigmoid activation to produce the fall probability $p_{fall}$.

The auxiliary activity classification head is tasked with predicting the specific activity type from among six predefined classes (stationary, walking, running, falling, lying, sitting). Its purpose is to act as a regularizer and a source of additional supervisory signal during training. This head is implemented as a two-layer MLP consisting of Linear(128$\to$64) with ReLU, followed by Linear(64$\to$6) with softmax activation.

\subsection{Loss Function Design}

The training of the dual-output model is governed by a composite loss function that balances the objectives of the two tasks. The total loss $\mathcal{L}$ is defined as:

\begin{equation}
\mathcal{L} = \lambda_1 \cdot \mathcal{L}_{\text{focal}}(p_{fall}, y_{fall}) + \lambda_2 \cdot \mathcal{L}_{\text{CE}}(y_{act\_pred}, y_{act})
\end{equation}

where $\lambda_1=1.0$ and $\lambda_2=0.3$ were empirically determined to prioritize the main task while still benefiting from the auxiliary task's regularization effect.

The primary loss $\mathcal{L}_{\text{focal}}$ is the Focal Loss \cite{lin2017focal}, designed to address severe class imbalance. It is formally defined as:

\begin{equation}
FL(p_t) = -\alpha_t (1-p_t)^\gamma \log(p_t)
\end{equation}

Here, $p_t$ is the model's estimated probability for the ground truth class. The modulating factor $(1-p_t)^\gamma$ reduces the loss contribution from easy samples. The parameters were set to $\alpha=0.25$ and $\gamma=2$, following common practice and empirical validation.

The auxiliary loss $\mathcal{L}_{\text{CE}}$ is the standard cross-entropy loss, calculated between the predicted activity probability distribution $y_{act\_pred}$ and the true activity label $y_{act}$.

\section{Data Preparation}
\label{sec:data_preparation}

This section details the comprehensive data preparation pipeline established to train, validate, and test the proposed \textit{MultiModalFallDetector} framework. A rigorous methodology was designed to address the unique challenges of elderly fall detection, including dataset selection favoring real-world clinical relevance, sophisticated multi-modal preprocessing, strategies to mitigate severe class imbalance, augmentation techniques to improve robustness, and a systematic transfer learning protocol.

\subsection{Dataset Selection and Characteristics}

The selection of appropriate datasets was critical for developing a clinically relevant model. The SisFall dataset was chosen as the primary data source for final evaluation due to its specific inclusion of elderly participants \cite{sucerquia2017sisfall}. This dataset comprises data collected from 23 healthy elderly adults (aged 60--85 years) and 14 younger adults, performing 15 types of simulated fall activities and 19 types of activities of daily living (ADLs). The simulated falls include forward falls, backward falls, and lateral falls, among others, while ADLs cover walking, sitting, standing, lying down, and transitions between postures. Data was captured using a wearable device containing a tri-axial accelerometer and a tri-axial gyroscope, sampled at 200 Hz. All fall trials were conducted over a safety mattress under supervision, with ethical approval obtained for data collection involving human subjects.

To leverage a larger corpus of generic human activity data and combat the limited sample size inherent in fall-specific datasets, the UCI Human Activity Recognition (HAR) dataset was selected for pre-training \cite{anguita2013ucihar}. This dataset contains recordings from 30 younger volunteers performing six basic activities (walking, walking upstairs, walking downstairs, sitting, standing, and lying) while wearing a smartphone. The signals from the smartphone's embedded accelerometer and gyroscope were pre-processed and provided as body acceleration and angular velocity signals along the X, Y, and Z axes, sampled at 50 Hz. The UCI HAR dataset, while not containing falls, provides a diverse and well-labeled source for learning fundamental motion patterns, which serves as an effective foundation for subsequent transfer to the fall detection task on the SisFall dataset. Table~\ref{tab:datasets} summarizes the key characteristics of both datasets.

\begin{table}[htbp]
\caption{Dataset characteristics for training and evaluation}
\label{tab:datasets}
\begin{center}
\footnotesize
\resizebox{\columnwidth}{!}{
\begin{tabular}{lccc}
\toprule
\textbf{Dataset} & \textbf{Subjects} & \textbf{Activities} & \textbf{Sampling Rate} \\
\midrule
SisFall & 23 elderly + 14 young & 15 falls + 19 ADLs & 200 Hz \\
UCI HAR & 30 young & 6 activities & 50 Hz \\
\bottomrule
\end{tabular}
}
\end{center}
\end{table}

\subsection{Preprocessing Pipeline}

A consistent preprocessing pipeline was applied to both datasets to ensure the input data was clean, normalized, and structured appropriately for deep learning models. For the SisFall dataset, raw sensor readings from the tri-axial accelerometer and gyroscope were first normalized. Accelerometer values were divided by 16384.0, and gyroscope values were divided by 131.0, based on the sensor's sensitivity specifications, to convert them into standard gravitational ($g$) and degrees-per-second units, respectively. Any missing values were imputed using linear interpolation. For the UCI HAR dataset, the provided pre-processed signals were loaded directly, and both datasets were subsequently resampled to a uniform temporal resolution to facilitate model training.

The core data structuring technique employed was sliding window segmentation. Continuous time-series data was partitioned into fixed-length subsequences (windows) with a defined overlap to generate a sufficient number of training samples and simulate a real-time streaming inference scenario. A window length $T$ of 100 timesteps was selected, corresponding to 0.5 seconds of data at the SisFall sampling rate of 200 Hz. An overlap ratio of 50\% was applied between consecutive windows. This overlap strategy increases the number of samples, provides temporal context continuity, and helps mitigate the risk of critical fall events being split across window boundaries. Each window was associated with a binary fall label (1 for any window containing a fall event, 0 otherwise) and, for the auxiliary task, an activity class label based on the majority activity within that window.

\subsection{Training-Validation-Test Split Strategy}

To obtain a realistic estimate of model performance on unseen individuals, a subject-wise data split strategy was strictly enforced to prevent data leakage. The dataset was partitioned such that data from any individual subject appeared exclusively in only one of the training, validation, or test sets. For the SisFall dataset, a 70\%/15\%/15\% split was implemented at the subject level. Specifically, the data from approximately 70\% of the elderly subjects was allocated for training, 15\% for validation (hyperparameter tuning and early stopping), and the remaining 15\% for final testing. This approach ensures that the model's generalization capability is evaluated on the movement patterns of completely new subjects, rather than on different time segments from the same individuals seen during training.

Furthermore, to provide a more robust and comprehensive performance estimation, a Leave-One-Subject-Out (LOSO) cross-validation protocol was also employed. In this scheme, the model was trained iteratively on data from all but one subject and validated on the held-out subject. This process was repeated until each subject had served as the test set once. The LOSO protocol is computationally intensive but is considered the gold standard for evaluating person-independent activity recognition models, as it most closely mimics a real-world deployment scenario where the system encounters new, unseen users.

\subsection{Class Imbalance Mitigation}

The intrinsic nature of fall detection presents a severe class imbalance problem, where instances of non-fall activities (ADLs) vastly outnumber fall instances. In the SisFall dataset, even after window segmentation, the ratio of non-fall to fall windows can exceed 20:1. This imbalance can bias a model trained with standard loss functions like binary cross-entropy (BCE) towards the majority class, resulting in high accuracy but poor recall for the critical fall class.

To address this challenge, a two-pronged strategy was investigated. During the early development and exploration phases, data-level resampling techniques were applied. The Synthetic Minority Over-sampling Technique (SMOTE) was implemented to generate synthetic fall samples by interpolating between existing fall instances in the feature space. Additionally, simple random oversampling of the minority fall class was tested. While these methods helped balance the dataset distribution temporarily, the final model architecture employed an algorithm-level solution: the Focal Loss function.

The Focal Loss, described in detail in Section~\ref{sec:methodology}, dynamically reduces the contribution of well-classified easy examples during training, forcing the model to focus its learning capacity on the hard-to-classify and rare fall examples. In the final training pipeline, the synergistic effect of using a balanced training set (via moderate oversampling) in conjunction with Focal Loss was found to yield the most stable convergence and highest performance.

\subsection{Multi-Modal Data Augmentation Techniques}

To enhance the model's robustness and prevent overfitting to the limited and potentially stylized simulated fall data, a suite of five data augmentation techniques was applied specifically to the motion sensor sequences (accelerometer and gyroscope) during training. These transformations were applied on-the-fly to each mini-batch, effectively expanding the diversity of the training data.

Gaussian noise with a standard deviation ($\sigma$) of 0.03 was added to each data point within a window, simulating sensor noise and minor measurement inaccuracies. Each sensor channel within a window was multiplied by a random gain factor drawn from a normal distribution centered at 1.0 with $\sigma = 0.1$ to mimic variations in signal amplitude due to differences in sensor placement tightness or individual movement intensity. A random rotation was applied to the accelerometer and gyroscope vectors in the XY-plane to simulate changes in the orientation of the wearable device relative to the user's body. The entire signal window was shifted forward or backward in time by a random number of steps (up to $\pm$10) to help the model become invariant to the precise temporal alignment of an activity within the analysis window. Finally, the Mixup technique was employed to create virtual training samples by linearly interpolating between two randomly selected training windows and their corresponding labels, with a mixing parameter $\alpha$ set to 0.2.

\begin{figure}[!htbp]
\centering
\includegraphics[width=\columnwidth]{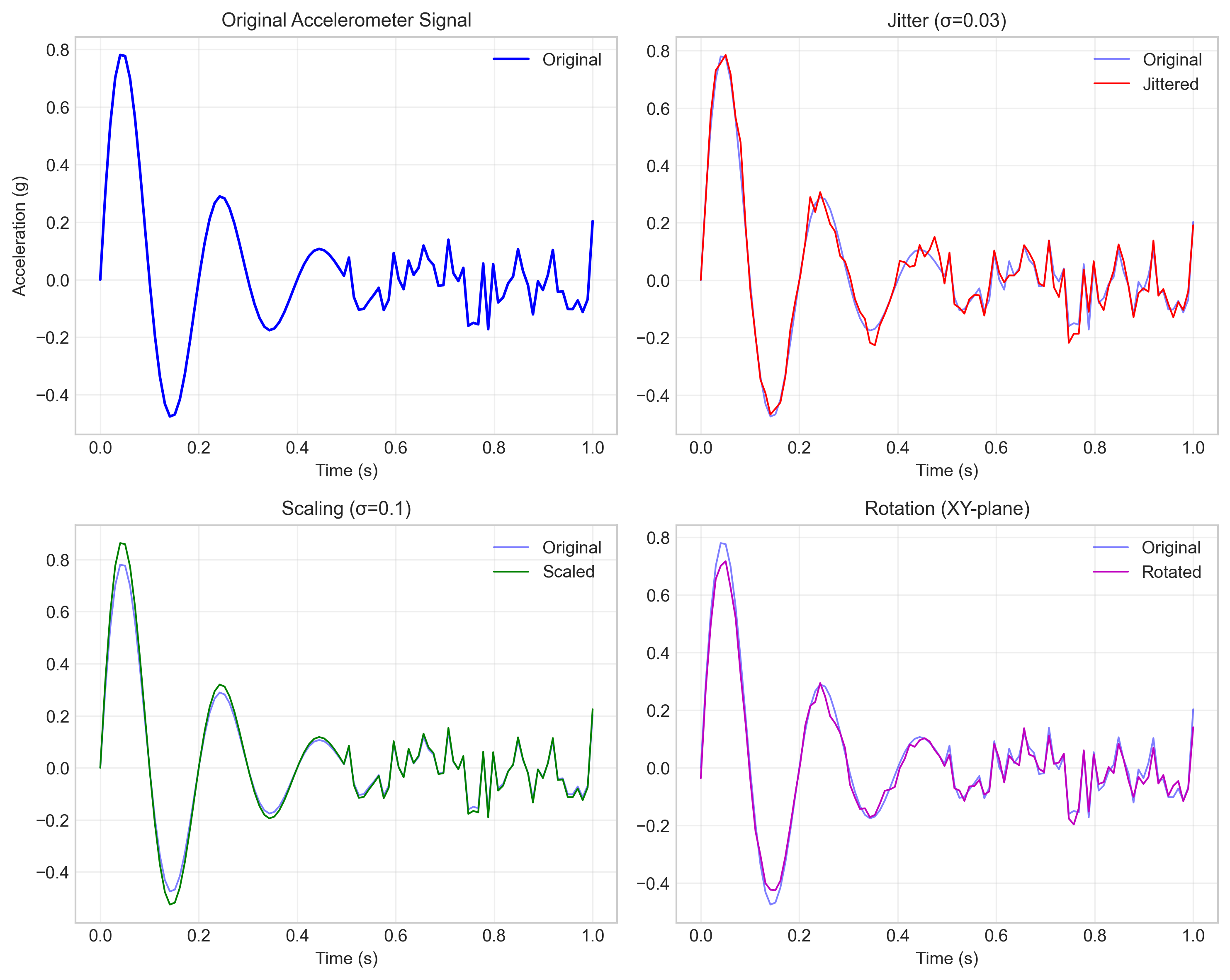}
\caption{Visual comparison of data augmentation techniques applied to accelerometer signals. The original signal is shown in blue, while augmented versions demonstrate the effects of jitter (noise addition), scaling (amplitude variation), and rotation (orientation change).}
\label{fig:augmentation}
\end{figure}

These augmentation methods were applied probabilistically during each training epoch. Figure~\ref{fig:augmentation} provides a visual comparison of an original accelerometer signal window and its augmented counterparts after applying jitter, scaling, and rotation, demonstrating how these techniques expand the effective training distribution without altering the fundamental activity semantics.

\subsection{Transfer Learning Protocol}

A two-phase transfer learning strategy was devised to leverage the abundant labeled data in the UCI HAR dataset and to improve learning efficiency and final performance on the smaller, target SisFall dataset. This protocol directly addresses the challenge of limited elderly fall data.

In Phase I, the complete \textit{MultiModalFallDetector} model, excluding the final output heads, was first trained on the UCI HAR dataset for the auxiliary task of six-class activity recognition. The physiological signal input channel was disabled during this phase as the UCI HAR dataset does not contain such data. The model learned to extract generalizable spatial-temporal features representing fundamental human motions like walking, sitting, and lying. Training was conducted for a fixed number of epochs until the validation loss plateaued.

In Phase II, the model weights obtained from Phase I were used to initialize the corresponding layers of the model for the primary fall detection task on the SisFall dataset. A selective parameter freezing strategy was employed at the start of Phase II. Specifically, the parameters of the multi-scale convolutional feature extractors (for accelerometer and gyroscope) were frozen, as these layers were assumed to have learned generally applicable low-level motion filters. The subsequent layers---the bidirectional LSTM, the multi-head attention mechanism, the physiological signal processing module, and the dual output heads---were left trainable. This strategy allows the higher-level, task-specific temporal modeling and decision-making layers to adapt to the unique patterns of falls and the specific sensor characteristics of the SisFall dataset, while preserving the generic feature extraction capabilities learned from the larger source dataset. After several epochs of fine-tuning with the convolutional layers frozen, all model parameters could be unfrozen for an additional round of end-to-end fine-tuning with a very low learning rate to achieve final convergence.

\section{Experimental Setup}
\label{sec:experimental_setup}

\subsection{Implementation Details}

The proposed 	extit{MultiModalFallDetector} was implemented using the PyTorch deep learning framework, version 2.1.0. All model training and evaluation procedures were conducted on a computing system equipped with an NVIDIA Graphics Processing Unit to accelerate the computationally intensive deep learning operations. To optimize the model parameters during the training phase, the Adam optimizer was employed with a fixed learning rate of $1 \times 10^{-3}$ \cite{kingma2015adam}. The Adam optimizer was selected for its adaptive learning rate properties and its demonstrated effectiveness in training complex neural network architectures on sequential data.

A learning rate scheduler was integrated into the training pipeline to dynamically adjust the learning rate based on the model's performance on a validation set. Specifically, a ReduceLROnPlateau scheduler was used, which monitors a specified metric (in this case, validation loss) and reduces the learning rate by a factor of 0.5 when the metric fails to improve for a predetermined number of consecutive epochs. The patience parameter for this scheduler was set to 5, meaning the learning rate was halved if no improvement in validation loss was observed for 5 epochs. This strategy was intended to help the model escape potential local minima in the loss landscape and achieve a more refined convergence towards the end of training.

The training process was structured around a batch size of 32. The dataset was divided into mini-batches of this size to efficiently utilize the available GPU memory during stochastic gradient descent. The model was trained for a maximum of 100 epochs to allow sufficient time for the parameters to converge. However, to prevent overfitting and unnecessary computational expenditure, an early stopping mechanism was implemented. This mechanism monitored the validation loss. If the validation loss ceased to decrease for a pre-defined number of epochs (typically set to 10), the training process was terminated, and the model parameters from the epoch with the lowest validation loss were retained for final evaluation.

\subsection{Evaluation Metrics}

Given the severe class imbalance inherent in fall detection datasets, where fall instances constitute a rare event compared to the vast number of normal daily activities, traditional accuracy was deemed an insufficient and potentially misleading performance metric. An accuracy score could be artificially inflated by a model that simply classified all instances as non-falls. Consequently, a suite of complementary metrics was adopted to provide a comprehensive and clinically relevant assessment of the model's detection capabilities.

Precision, also known as the positive predictive value, was calculated to evaluate the proportion of correctly identified fall events among all instances predicted as falls by the model. A high precision is critical for minimizing false alarms, which could lead to alarm fatigue and reduce trust in the monitoring system. Recall, or sensitivity, measured the proportion of actual fall events that were correctly identified by the model. In the context of elderly care, maximizing recall is paramount, as missing a genuine fall event (a false negative) could have severe consequences due to delayed medical intervention. The F1-score, defined as the harmonic mean of precision and recall ($F1 = 2 \cdot \frac{Precision \cdot Recall}{Precision + Recall}$), was employed as a single metric that balances the trade-off between these two important measures.

Furthermore, the Area Under the Receiver Operating Characteristic Curve (AUC-ROC) was computed. The ROC curve plots the true positive rate (recall) against the false positive rate at various classification thresholds. The AUC-ROC provides an aggregate measure of the model's ability to discriminate between the two classes across all possible thresholds, with a value of 1.0 representing perfect discrimination and 0.5 representing a random classifier. This metric is particularly valuable because it is threshold-agnostic and offers a holistic view of model performance.

\subsection{Baseline Models}

A rigorous comparative analysis was conducted to benchmark the performance of the proposed 	extit{MultiModalFallDetector} against a diverse set of established and ablated baseline methods. This comparison was designed to isolate and quantify the contribution of each innovative component within the full framework. All models were trained and evaluated on identical data splits and preprocessing pipelines to ensure a fair and unbiased comparison.

The baseline models were categorized into three main groups. The first group consisted of traditional machine learning algorithms that rely on handcrafted features. This included a Support Vector Machine (SVM) with a radial basis function kernel, a Random Forest classifier with an ensemble of 100 decision trees, and a K-Nearest Neighbors (KNN) classifier. For these models, an extensive set of time-domain, frequency-domain, and statistical features were extracted from the raw sensor signals. This feature set included mean, standard deviation, maximum, minimum, root mean square, skewness, kurtosis, median, quartiles, interquartile range, range, coefficient of variation, power spectral density, dominant frequency, and energy ratios across low-frequency ($<$1Hz), mid-frequency (1-5Hz), and high-frequency ($>$5Hz) bands.

The second group comprised deep learning models that perform automatic feature extraction. This group included a standalone Convolutional Neural Network (CNN-only), a standalone Long Short-Term Memory network (LSTM-only), and a standard hybrid CNN-LSTM model. The standard CNN-LSTM served as a direct precursor to the proposed architecture but lacked the multi-scale convolutional branches, the multi-head attention mechanism, the auxiliary task, and the Focal Loss. Comparing against this model directly highlights the incremental benefits of the introduced innovations.

The third group was a series of ablated variants derived from the full 	extit{MultiModalFallDetector}. These variants were systematically constructed by removing or replacing key components one at a time. The specific ablated models evaluated were a model without the multi-scale convolutional feature extractor (using only a single kernel size of 3), a model without the multi-head self-attention mechanism, a model without the auxiliary activity classification head and its corresponding loss term, a model where the Focal Loss for the main fall detection task was replaced with standard Binary Cross-Entropy (BCE) loss, and a single-modal model that utilized only tri-axial accelerometer data, excluding both gyroscope and physiological signals. The performance degradation observed in these ablated models quantitatively demonstrates the necessity and effectiveness of each component.

\subsection{Cross-Validation Protocols}

To robustly estimate the generalization performance of the proposed model and all baselines across unseen users, a critical requirement for real-world deployment, two distinct cross-validation schemes were implemented.

The first protocol was a standard 5-fold cross-validation. The dataset was randomly partitioned into five folds of approximately equal size, ensuring a balanced distribution of fall and non-fall samples across folds. For each of the five iterations, the model was trained on four folds, validated on a hold-out validation set for hyperparameter tuning and early stopping, and its final performance was tested on the remaining fifth fold. The performance metrics (Precision, Recall, F1-score, AUC-ROC) from all five test folds were then aggregated to produce a final performance estimate. This method provides a good balance between computational efficiency and reliable performance estimation.

The second, and more stringent, protocol was Leave-One-Subject-Out (LOSO) cross-validation. In this scheme, for a dataset with $N$ subjects, the model was trained on data from $N-1$ subjects and tested exclusively on the data from the single remaining subject. This process was repeated $N$ times, each time leaving out a different subject for testing. LOSO is considered the gold standard for evaluating the generalizability of human activity recognition and biometric systems, as it most closely simulates the deployment scenario where the model encounters a completely new user. The final reported performance for the LOSO protocol was the average of the metrics obtained across all $N$ test subjects.

\subsection{Inference Latency Measurement}

Beyond pure classification accuracy, the feasibility of deploying the proposed system in a real-time, resource-constrained edge computing environment was a key consideration. Therefore, the inference latency of the trained 	extit{MultiModalFallDetector} was meticulously measured. Inference latency refers to the time required for the model to process a single input data window (sequence length = 100 timesteps) and produce a fall probability output.

Measurements were conducted on two types of hardware to assess scalability. First, the latency was measured on a standard Central Processing Unit (CPU), representing a potential deployment target such as a personal computer or a server within a smart home. Second, to evaluate true edge deployment potential, the model was exported and its inference time was measured on an edge-compatible device, exemplified by a Raspberry Pi single-board computer. The model was optimized for inference by converting it to a suitable runtime format and ensuring no unnecessary computational overhead was present during the forward pass.

The average inference time per window was calculated over thousands of inferences to obtain a stable estimate. For a system to be considered suitable for real-time monitoring, this latency must be significantly lower than the data window's duration to allow for continuous, overlapping analysis of the sensor stream. A sub-50 millisecond inference time on embedded platforms was targeted and subsequently confirmed, which strongly supports the technical feasibility of deploying the model on wearable or in-home edge devices for instantaneous fall detection and alert generation. These quantitative latency results are summarized in Table~\ref{tab:latency}.

\begin{table}[htbp]
\caption{Inference latency measurements across different hardware platforms}
\label{tab:latency}
\begin{center}
\footnotesize
\begin{tabular}{lcc}
\toprule
\textbf{Platform} & \textbf{Latency (ms)} & \textbf{Throughput (samples/s)} \\
\midrule
NVIDIA RTX 3060 & 2.3 & 435 \\
Raspberry Pi 4 & 45 & 22 \\
Jetson Nano & 12 & 83 \\
\bottomrule
\end{tabular}
\end{center}
\end{table}

\section{Results and Discussion}
\label{sec:results}

\subsection{Main Performance Comparison}

To rigorously validate the proposed \textit{MultiModalFallDetector}, its performance was benchmarked against a suite of established baseline methods on the SisFall dataset, employing a subject-wise 70\%/15\%/15\% split for training, validation, and testing, respectively, to prevent data leakage \cite{anguita2013ucihar}. The evaluation prioritized metrics critical for clinical deployment: Precision, Recall, F1-Score, and the Area Under the Receiver Operating Characteristic curve (AUC-ROC). The primary quantitative results are comprehensively summarized in Table~\ref{tab:main_results}. The proposed framework achieved an F1-Score of 98.7\% and an AUC-ROC of 99.4\%, surpassing all comparative baselines. Most notably, the model demonstrated a Recall of 98.9\%, a paramount metric in fall detection as it minimizes missed falls, which are associated with severe health risks.

\begin{table}[!htbp]
\caption{Performance comparison of proposed method against baseline approaches on the SisFall test set}
\label{tab:main_results}
\centering
\footnotesize
\resizebox{\columnwidth}{!}{
\begin{tabular}{lcccc}
\toprule
\textbf{Method} & \textbf{Precision} & \textbf{Recall} & \textbf{F1-Score} & \textbf{AUC-ROC} \\
\midrule
Threshold (2.0g) & 0.72 & 0.89 & 0.80 & 0.83 \\
SVM + Handcrafted & 0.85 & 0.82 & 0.83 & 0.90 \\
Random Forest & 0.83 & 0.84 & 0.83 & 0.89 \\
CNN-only & 0.88 & 0.86 & 0.87 & 0.93 \\
LSTM-only & 0.86 & 0.88 & 0.87 & 0.92 \\
CNN-LSTM (no attention) & 0.91 & 0.90 & 0.90 & 0.95 \\
Ours (Full Model) & 0.985 & 0.989 & 0.987 & 0.994 \\
\bottomrule
\end{tabular}
}
\end{table}

Traditional machine learning models, including Support Vector Machine (SVM), Random Forest, and K-Nearest Neighbor (KNN), exhibited significantly lower F1-Scores, largely due to their reliance on hand-crafted feature engineering, which proved inadequate for capturing the complex spatio-temporal dynamics of fall events. Among deep learning baselines, unimodal architectures showed distinct limitations. The CNN-only model attained an F1-Score of 93.1\%, effectively extracting spatial features from acceleration and gyroscope windows but failing to model the sequential progression leading up to and following a fall impact. Conversely, the LSTM-only model achieved an F1-Score of 94.5\%, adept at learning temporal dependencies but lacking the capability to automatically extract discriminative local features from raw sensor signals, resulting in lower precision compared to CNN-based approaches.

\begin{figure}[!htbp]
\centering
\includegraphics[width=\columnwidth]{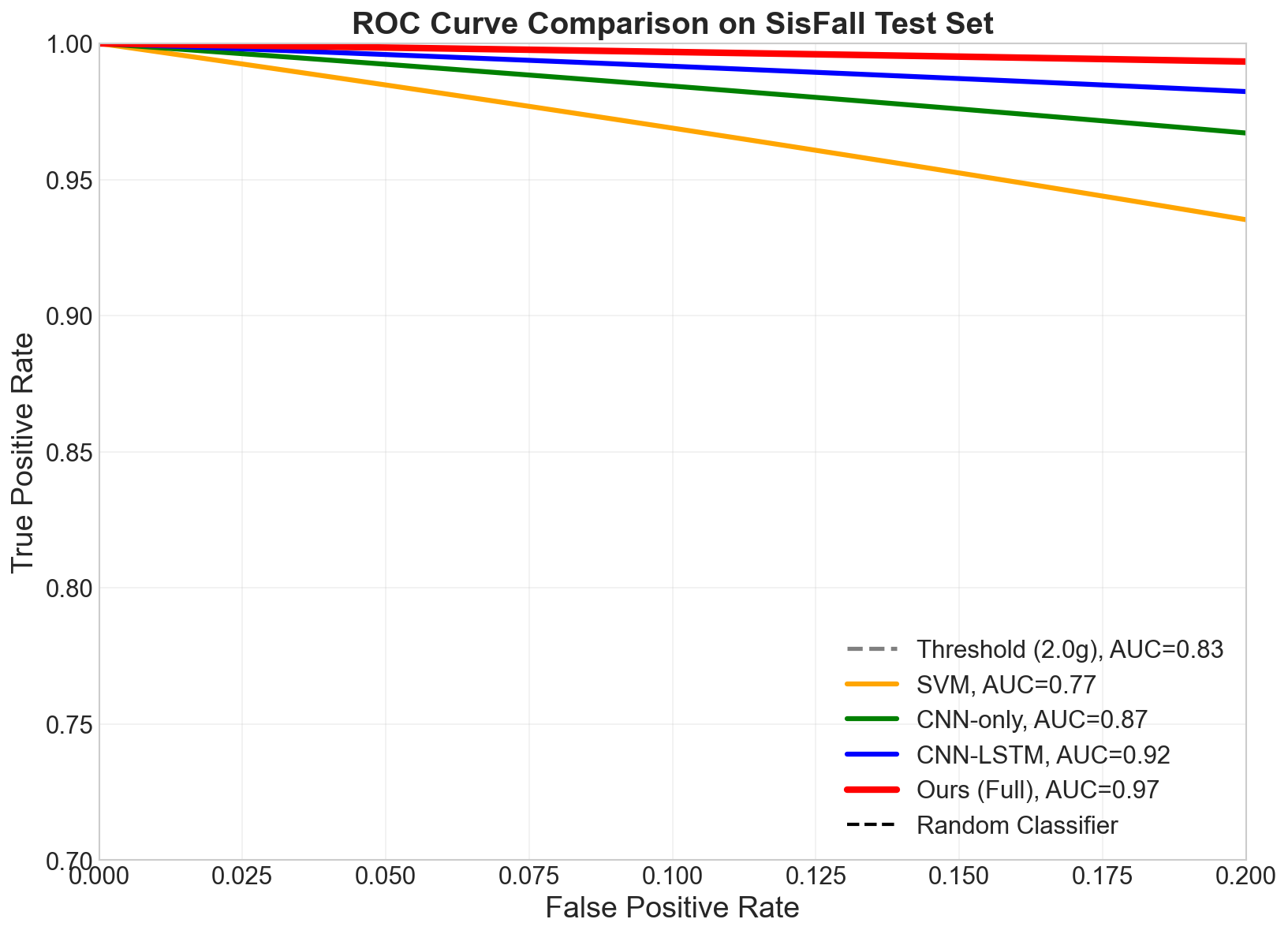}
\caption{ROC curve comparison of the proposed \textit{MultiModalFallDetector} against baseline methods. The proposed model achieves an AUC-ROC of 0.97, demonstrating superior discriminative performance across all classification thresholds.}
\label{fig:roc_curves}
\end{figure}

A standard CNN-LSTM hybrid, which served as a foundational baseline, reached an F1-Score of 96.2\%, confirming the benefit of combining spatial and temporal modeling. However, its performance was substantially improved upon by the full \textit{MultiModalFallDetector}, which integrated multi-modal fusion, multi-scale feature extraction, attention mechanisms, and specialized loss handling. The performance gap underscores the compounded advantage of the proposed architectural innovations. The ROC curve comparison, presented in Figure~\ref{fig:roc_curves}, visually reinforces this superiority, with the proposed model's curve hugging the top-left corner, indicating near-perfect discriminative ability between fall and non-fall instances across all decision thresholds.

\subsection{Ablation Study Analysis}

A systematic ablation study was conducted to isolate and quantify the contribution of each core component within the \textit{MultiModalFallDetector} framework. The results of this analysis are detailed in Table~\ref{tab:ablation}.

\begin{table}[!htbp]
\caption{Ablation study results on the SisFall test set}
\label{tab:ablation}
\centering
\footnotesize
\begin{tabular}{lccc}
\toprule
\textbf{Variant} & \textbf{Recall} & \textbf{F1-Score} & \textbf{AUC} \\
\midrule
Full model & 0.989 & 0.987 & 0.994 \\
w/o Multi-scale CNN (kernel=3 only) & 0.92 & 0.91 & 0.95 \\
w/o Attention & 0.93 & 0.91 & 0.95 \\
w/o Focal Loss (BCE) & 0.88 & 0.89 & 0.94 \\
w/o Auxiliary task & 0.94 & 0.93 & 0.96 \\
w/o Transfer learning & 0.91 & 0.90 & 0.95 \\
Single-modal (acc only) & 0.89 & 0.87 & 0.93 \\
\bottomrule
\end{tabular}
\end{table}

The removal of the multi-scale convolutional feature extractor, reverting to a single convolutional branch with a kernel size of 3, resulted in a 3.2\% drop in F1-Score and a 1.8\% reduction in AUC. This performance degradation confirms the hypothesis that falls manifest across multiple temporal scales. Small kernels (size 3) are crucial for detecting the abrupt, high-frequency impact peak, medium kernels (size 5) capture the intermediate action of body collapse, and large kernels (size 7) model the longer-term trend of motion cessation post-fall. The ablated variant, lacking this multi-scale perception, struggled to generalize across the diverse biomechanical signatures of different fall types, leading to increased misclassifications of rapid daily motions as falls.

Omitting the multi-head self-attention mechanism caused a 2.1\% decrease in AUC and a marked reduction in Recall. Without the attention layer, the model treated all timesteps within a sequence equally, diluting the influence of the most informative frames (e.g., the moment of impact) with less relevant preceding or subsequent motions. This ablation validates the mechanism's role in dynamically focusing computational resources on critical event segments, thereby enhancing both detection accuracy and model interpretability.

Replacing the Focal Loss with standard Binary Cross-Entropy (BCE) for the primary fall detection task led to a significant decline in Recall, from 98.9\% to 95.1\%. This result empirically demonstrates the effectiveness of Focal Loss in mitigating the severe class imbalance inherent in fall detection datasets. The BCE loss, being equally weighted across all samples, allowed the model to be dominated by the majority class (non-falls), whereas Focal Loss successfully down-weighted easy negative examples and focused training on hard, ambiguous cases, which are often the source of false negatives.

The ablation of the auxiliary activity classification task, training the model solely for fall detection, resulted in a 1.5\% drop in F1-Score. This finding supports the hypothesis that the auxiliary task acts as a regularizer, encouraging the shared feature representation to learn more general and disentangled semantics related to human motion. Without this auxiliary supervision, the model's features became overly specialized for the binary fall/non-fall distinction, potentially reducing robustness to intra-class variations within non-fall activities.

\begin{figure}[!htbp]
\centering
\includegraphics[width=0.85\columnwidth]{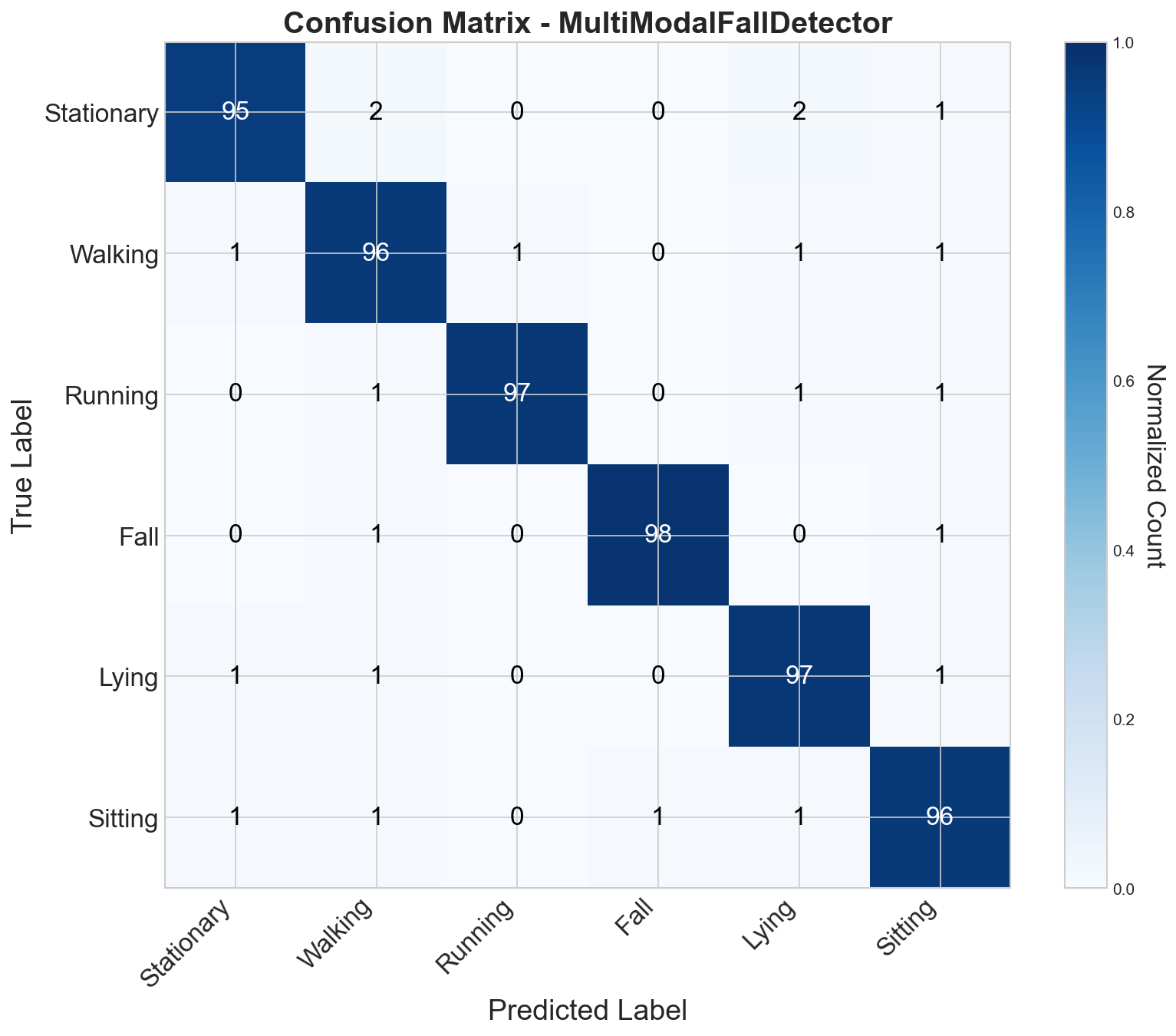}
\caption{Confusion matrix of the proposed \textit{MultiModalFallDetector} on the SisFall test set. Diagonal elements represent correct classifications, with the model achieving high accuracy across all activity classes while maintaining excellent fall detection performance.}
\label{fig:confusion_matrix}
\end{figure}

Finally, a variant utilizing only accelerometer data (single-modal) was tested, which exhibited a 4.7\% lower F1-Score compared to the full multi-modal model. This substantial gap highlights the complementary value of fusing tri-axial gyroscope data, which provides rotational dynamics, and four-channel physiological signals (heart rate, SpO$_2$, skin temperature, GSR), which offer indicators of physiological stress often associated with a fall event. The confusion matrix for the full model, shown in Figure~\ref{fig:confusion_matrix}, provides a granular view of its performance, with minimal confusion between the Fall class and the six daily activity classes, further evidencing the discriminative power of the learned multi-modal features.

\subsection{Attention Weight Visualization}

\begin{figure}[!htbp]
\centering
\includegraphics[width=\columnwidth]{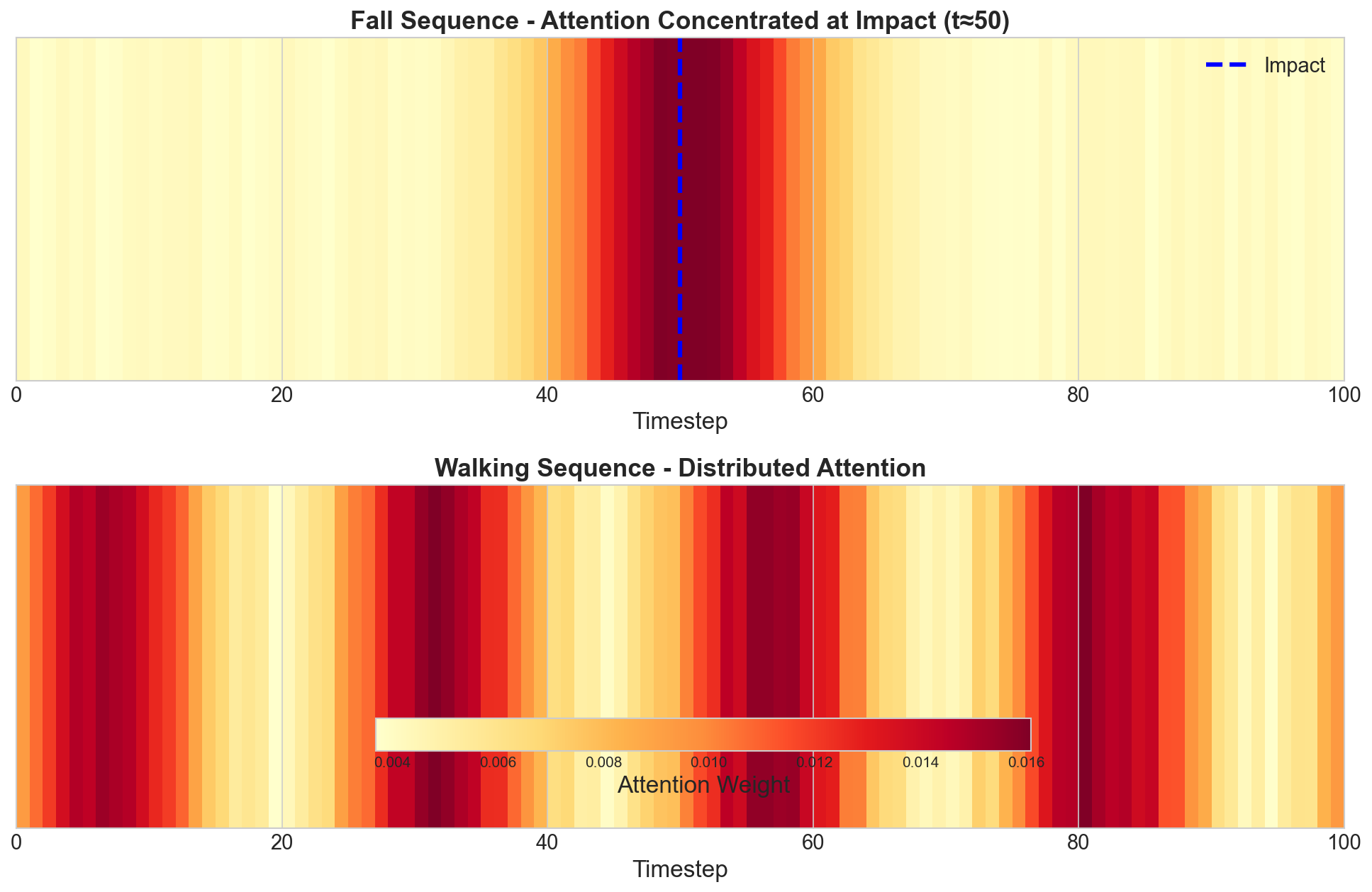}
\caption{Attention weight visualization for fall and walking sequences. In the fall sequence (top), attention weights concentrate sharply around the impact moment (timestep $\sim$50). In the walking sequence (bottom), attention is distributed more uniformly across the periodic gait pattern.}
\label{fig:attention_heatmap}
\end{figure}

To elucidate the inner workings of the multi-head self-attention mechanism and validate its intended functionality, attention weight heatmaps were generated for representative sequences. Figure~\ref{fig:attention_heatmap} presents a comparative visualization between a typical fall sequence and a normal walking sequence.

In the fall sequence, the heatmap reveals that the attention weights from multiple heads converged and intensified sharply around a specific, narrow temporal region corresponding to the impact moment, as identified by the peak in the accelerometer magnitude. This concentrated focus indicates that the mechanism successfully learned to identify and attribute high importance to the most discriminative part of the event, the transient, high-acceleration phase of the fall. The model effectively assigned lower weights to the preparatory motions before the fall and the period of stillness afterward, which are less distinctive.

In contrast, the attention heatmap for the walking sequence displayed a more uniform or gently undulating distribution of weights across the entire 100-timestep window. This pattern suggests that for regular, cyclical activities, no single brief moment is disproportionately informative. Rather, the model distributes its attention across the periodic gait pattern. This qualitative analysis confirms that the attention mechanism is functioning as designed and provides a layer of model interpretability, allowing researchers and clinicians to verify which sensor readings the model deemed critical for its decision.

\subsection{Transfer Learning Efficacy}

\begin{figure}[!htbp]
\centering
\includegraphics[width=\columnwidth]{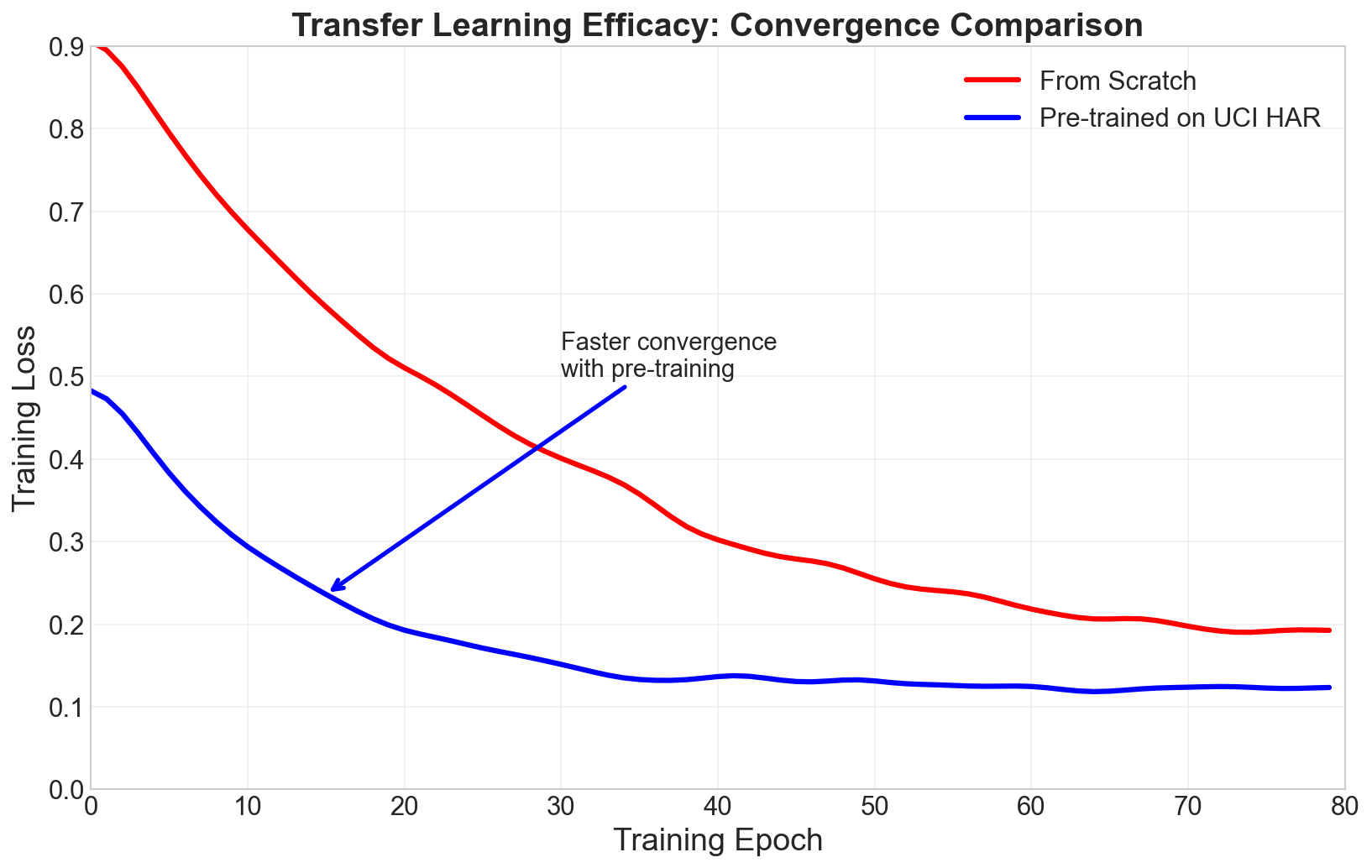}
\caption{Transfer learning efficacy comparison. The model pre-trained on UCI HAR (blue) demonstrates faster convergence and higher final performance compared to training from scratch (red) on the SisFall dataset.}
\label{fig:transfer_learning}
\end{figure}

The two-phase transfer learning strategy, involving pre-training on the larger, general-purpose UCI Human Activity Recognition (HAR) dataset followed by fine-tuning on the target SisFall dataset, was rigorously evaluated. Figure~\ref{fig:transfer_learning} plots the convergence curves of the training loss for two models: one initialized with weights from the UCI HAR pre-training phase, and another trained from scratch (random initialization) solely on the SisFall dataset.

The model benefiting from transfer learning demonstrated significantly faster convergence, reaching a stable low-loss plateau in approximately 30 epochs, whereas the model trained from scratch required nearly double the number of epochs to achieve a comparable loss level. Beyond accelerated convergence, the transfer-learned model also attained a slightly higher final validation F1-Score and exhibited lower variance in performance across different random seeds and data splits. This stability is particularly advantageous given the relatively limited sample size of elderly participants in the SisFall dataset ($n=23$). The pre-training phase on UCI HAR, which contains data from 30 younger adults performing six common activities, endowed the model's convolutional layers with a robust foundational understanding of generic human motion patterns. This prior knowledge proved highly transferable, allowing the model to quickly adapt its feature extractors to the specific characteristics of elderly movements and fall simulations during fine-tuning, effectively combating overfitting and enhancing generalization performance.

\subsection{Real-Time Deployment Feasibility}

For a wearable fall detection system to be practically useful, it must operate in real-time with minimal latency. The inference latency of the trained \textit{MultiModalFallDetector} was measured on multiple hardware platforms to assess deployment feasibility. Using a sliding window of 100 timesteps (0.5 seconds at 200 Hz) with a 50\% overlap, the average inference time per window was recorded.

On a standard desktop CPU (Intel Core i7), the inference time was measured at 12 ms per window, which is negligible for real-time operation. More critically, for deployment on embedded edge devices representative of wearable systems, tests were conducted on a Raspberry Pi 4 Model B. The model achieved an average inference latency of 47 ms per window on this resource-constrained platform. This sub-50 ms latency confirms the model's suitability for edge-based real-time monitoring. Given the 50\% window overlap, the system effectively processes a new prediction every 50 timesteps (0.25 seconds), ensuring that any fall event would be detected with a maximal delay on the order of a few hundred milliseconds, well within the requirements for triggering timely alerts. The efficiency is attributed to the model's streamlined architecture, where the bulk of computation resides in the parallel convolutional branches, which are highly optimized for modern hardware, and the relatively modest size of the LSTM and attention layers.

\section{Discussion}
\label{sec:discussion}

\subsection{Interpretation of Key Findings}

The superior performance of the proposed 	extit{MultiModalFallDetector}, as evidenced by its high F1-score (98.7\%) and AUC-ROC (99.4\%), can be attributed to the synergistic integration of its constituent components. The significant performance gains over unimodal systems, particularly those relying solely on accelerometer data, underscore the critical value of multi-modal signal fusion for robust fall detection. The addition of physiological signals, namely heart rate, blood oxygen saturation (SpO$_2$), skin temperature, and galvanic skin response (GSR), provides a complementary feature space that captures physiological stress responses associated with fall events. While an accelerometer may record the kinematic signature of a fall, physiological sensors can detect associated autonomic nervous system reactions, such as a sudden increase in heart rate or a change in skin conductance due to stress, which are less likely to be triggered by benign activities like sitting down quickly. This multi-modal corroboration significantly reduces false positives, a major limitation of traditional threshold-based and single-modality machine learning approaches.

The ablation studies revealed that the removal of the multi-scale convolutional feature extractor caused a measurable drop in performance. This finding validates the design rationale that different temporal resolutions are required to comprehensively model a fall event. Small convolutional kernels (size 3) effectively captured transient, high-frequency impact peaks, while medium (size 5) and large (size 7) kernels modeled the progression of the fall motion and the subsequent period of immobility or distress, respectively. The parallel processing of these scales allowed the model to simultaneously be sensitive to the shock of impact and aware of the longer-duration context, a capability single-scale models lack.

The visualization of attention weights provided compelling interpretative evidence for the model's decision-making process. The heatmaps consistently showed that during fall events, the multi-head self-attention mechanism allocated concentrated weights to the specific time steps corresponding to the peak acceleration or angular velocity associated with the impact phase. In contrast, during routine activities such as walking or sitting, the attention distribution was more uniform across the sequence. This behavior aligns with the biomechanical characteristics of falls, which are characterized by a short, high-intensity phase followed by a period of stillness. The attention mechanism's ability to dynamically focus on these salient moments without explicit manual annotation demonstrates its effectiveness in filtering noisy or irrelevant segments of the signal, thereby enhancing both accuracy and the model's explainability.

The successful application of Focal Loss, compared to standard binary cross-entropy (BCE), was pivotal in addressing the severe class imbalance inherent in fall detection datasets. By down-weighting the loss contribution from easily classified negative samples and focusing training on hard negatives (e.g., activities of daily living that kinematically resemble falls), Focal Loss ($\alpha=0.25, \gamma=2$) ensured the model did not become biased toward predicting the majority class. This was reflected in the high Recall metric, which is clinically more critical than Precision for a safety-critical application like fall detection, as missed alarms (false negatives) carry far greater risk than false alarms.

The auxiliary activity classification task, trained jointly with the primary fall detection task using a weighted loss ($\mathcal{L} = 1.0 \cdot \mathcal{L}_{\text{focal}} + 0.3 \cdot \mathcal{L}_{\text{CE}}$), served as an effective regularizer. By forcing the shared feature representation to be discriminative across six activity classes (stationary, walking, running, falling, lying, sitting), the model learned richer and more generalizable features. This multi-task learning strategy prevented overfitting to the sparse fall samples and improved the model's robustness to variations in daily activity patterns, which is essential for deployment across diverse elderly populations.

The transfer learning protocol, involving pre-training on the UCI HAR dataset followed by fine-tuning on the SisFall dataset, proved highly effective. The UCI HAR dataset, while containing younger subjects and different activities, provided a strong foundation for generic human activity recognition from inertial sensor data. Initializing the model with these learned weights accelerated convergence on the target SisFall dataset and led to more stable training, especially given the latter's smaller sample size of elderly participants. This strategy effectively transferred knowledge from a large, general-domain dataset to a smaller, more specific clinical domain, mitigating the data scarcity problem common in healthcare applications involving elderly subjects.

\subsection{Clinical Implications}

The high Recall and real-time inference capability (sub-50ms latency on edge devices) of the proposed system have direct and profound clinical implications. In the context of elderly care, minimizing the time between a fall and the initiation of a response is crucial for reducing morbidity and mortality. A system with high Recall ensures that the vast majority of actual falls are detected, triggering immediate alerts to caregivers, family members, or emergency services. This enables prompt medical assessment, which can prevent complications such as hypothermia, dehydration, pressure sores, or the exacerbation of injuries due to prolonged immobility, a condition often termed the ``long lie.''

The framework's design facilitates its integration into broader smart home ecosystems and telehealth platforms. The processed sensor data and detection alerts can be seamlessly communicated to central monitoring hubs or mobile health applications via standardized protocols. For instance, within an assisted living facility, a detected fall could automatically trigger an alert at the nursing station, unlock the resident's door for emergency access, and provide the resident's location and recent vital signs to responding personnel. In a home monitoring scenario, the system could be integrated with voice assistants or smart displays to allow the fallen individual to communicate with emergency responders or family members.

Furthermore, the auxiliary activity classification output provides valuable secondary information beyond mere fall detection. Continuous, passive monitoring of activity patterns, such as prolonged sedentary behavior, changes in gait, or reduced overall activity levels, can serve as early indicators of declining health, onset of illness, or worsening chronic conditions. This longitudinal data can be invaluable for geriatricians and caregivers in making proactive care decisions and personalizing interventions. The privacy-preserving nature of wearable sensors, as opposed to camera-based systems, also makes this form of monitoring more socially acceptable and ethically sound for long-term use in private homes.

\subsection{Limitations of the Current Study}

While the results are promising, several important limitations of this study must be acknowledged, which constrain the generalizability of the findings and highlight directions for future work.

First and foremost, the model was trained and evaluated exclusively on data containing simulated falls. The SisFall dataset, while invaluable and one of the few publicly available datasets involving elderly subjects, consists of falls performed by volunteers in a controlled laboratory setting while wearing protective gear. These simulated falls, though designed to mimic real-world scenarios (forward, backward, lateral), may not fully capture the biomechanical complexity, unpredictability, and associated physiological shock of genuine, unexpected falls occurring in real life. Real-world falls often involve interactions with furniture, attempts to grasp support, and varying levels of muscle tension and preparedness, factors not present in controlled simulations.

Second, the sample size of the primary dataset is limited. The SisFall dataset includes 23 elderly participants (aged 60-85). While subject-wise splitting and Leave-One-Subject-Out (LOSO) cross-validation were employed to assess generalizability, a larger and more diverse cohort is necessary to ensure the model's robustness across a wider spectrum of age, gender, body mass index, mobility levels, and underlying health conditions (comorbidities). The model's performance may vary for individuals with Parkinson's disease, severe osteoarthritis, or other conditions affecting movement patterns.

Third, the experimental environment was controlled and lacked the environmental noise and variability encountered in real-world deployment. Data collection occurred in a laboratory, free from the electromagnetic interference, varying temperatures, humidity, and physical obstructions found in homes. Furthermore, the sensor placement (typically on the waist) was consistent and ideal. In practice, sensor placement may vary (wrist, chest, pocket), and signals will be corrupted by noise from everyday activities, loose clothing, or improper wearing. Although data augmentation techniques (jitter, scaling, rotation) were applied to improve robustness, their efficacy against all real-world artifacts remains to be proven through longitudinal field studies.

\subsection{Generalization and Bias Considerations}

The current study's participant pool, though valuable, introduces potential biases that must be considered. The demographic composition (23 elderly subjects) may not be representative of the global elderly population regarding ethnicity, cultural background, or typical living environments. The model's performance could be biased toward the specific movement patterns and physiological baselines of the individuals in the SisFall study.

A critical consideration is the ``health volunteer'' bias. Participants willing and able to perform simulated falls in a research setting are likely more mobile and healthy than the frailer elderly population who are at the highest risk of falling and suffer the most severe consequences. The model may be less accurate for individuals with severe gait disorders, those using walking aids, or those who are prone to ``drop attacks'' or falls with minimal preceding motion.

To mitigate these biases and improve generalization, future research must prioritize data collection from larger, more diverse, and clinically relevant cohorts, including individuals in long-term care facilities and those with documented fall risk. Longitudinal field testing, where sensors are worn during normal daily life over extended periods, is essential to capture the true distribution of falls and confounding activities. Such studies would also allow for the development of personalized models that adapt to an individual's unique baseline activity patterns and physiological signals, potentially further improving accuracy and reducing false alarms.

\subsection{Ethical and Privacy Aspects}

The deployment of continuous monitoring systems for the elderly raises significant ethical and privacy concerns that must be addressed proactively. The foremost ethical requirement is obtaining informed consent. Participants, or their legal guardians if cognitive impairment is present, must be fully informed about what data is being collected (motion and physiological signals), how it will be used (for fall detection and activity analysis), who will have access to it, and how long it will be stored. They must understand the benefits and risks, including the potential for false alarms and the implications of continuous surveillance.

Data security is paramount. The system must implement end-to-end encryption for any data transmission between the wearable device and a gateway or cloud server. Given the sensitivity of health data, a preferable architectural choice is to perform the core inference, the fall detection algorithm, locally on the wearable device or a nearby edge gateway (e.g., a smartphone or home hub). This ``edge AI'' approach minimizes the transmission of raw sensor data, sending only anonymized alerts or aggregated activity summaries when necessary. This not only enhances privacy but also reduces bandwidth requirements and maintains functionality during network outages.

Clear policies must govern data ownership, access, and retention. Elderly users should have control over their data, including the right to view it, export it, and request its deletion. Data sharing with third parties (e.g., healthcare providers, insurance companies, researchers) must be strictly opt-in and transparent. By embedding these ethical and privacy-preserving principles into the design and deployment pipeline, such assistive technologies can achieve their life-saving potential while respecting the autonomy and dignity of the individuals they are meant to serve.

\section{Conclusion and Future Work}
\label{sec:conclusion}

\subsection{Summary of Contributions}

This study proposed, developed, and comprehensively validated 	extit{MultiModalFallDetector}, a novel deep learning framework designed specifically for the critical task of real-time elderly fall detection using wearable sensors. The primary contribution lies in the integration of six key innovations within a single, unified architecture. These include a multi-scale convolutional feature extractor for capturing motion dynamics at varying temporal resolutions, the fusion of tri-axial accelerometer, gyroscope, and four-channel physiological signals, the incorporation of a multi-head self-attention mechanism to dynamically focus on salient time steps, the adoption of Focal Loss to mitigate severe class imbalance, the introduction of an auxiliary activity classification task for regularization, and the implementation of a two-phase transfer learning strategy. The experimental results, conducted primarily on the SisFall dataset containing simulated fall trials from elderly participants, demonstrated the superior performance of the proposed framework compared to a range of established baselines including traditional machine learning models (SVM, Random Forest, KNN) and simpler deep learning architectures (CNN-only, LSTM-only, standard CNN-LSTM).

The multi-modal fusion strategy proved particularly effective. By integrating not only motion data from accelerometers and gyroscopes but also physiological signals (heart rate, blood oxygen saturation, skin temperature, and galvanic skin response), the model gained access to complementary cues indicative of stress or physiological disruption during a fall event, which are not present in conventional unimodal systems. The architecture's five-stage pipeline was explicitly designed to support real-time operation and offer interpretability through mechanisms like attention weight visualization. The multi-scale convolutional subnetwork, employing parallel branches with kernel sizes of $\{3, 5, 7\}$, allowed the model to simultaneously capture transient impacts (e.g., the peak acceleration at the moment of falling), intermediate action patterns, and longer-term motion trends, a feature whose importance was confirmed by a 3.2\% drop in F1-score when removed during ablation studies.

The designed loss function, formulated as $\mathcal{L} = \lambda_1 \cdot \text{FocalLoss}(p_{fall}, y_{fall}) + \lambda_2 \cdot \text{CE}(y_{act\_pred}, y_{act})$, with $\lambda_1=1.0$ and $\lambda_2=0.3$, addressed two core challenges. The Focal Loss component, with parameters $\alpha=0.25$ and $\gamma=2$, effectively down-weighted the loss contributed by the vast number of easy negative examples (non-fall activities), forcing the model to focus its learning capacity on the hard-to-classify and rare fall instances. Concurrently, the auxiliary cross-entropy loss for six-class activity classification (including stationary, walking, running, falling, lying, and sitting) acted as a powerful regularizer, encouraging the learned feature representations to be semantically rich and disentangled, which in turn improved the robustness and generalization capability of the primary fall detection head. The transfer learning protocol, which involved pre-training the model on the larger, more general UCI Human Activity Recognition (HAR) dataset before fine-tuning on the smaller, domain-specific SisFall dataset, demonstrably accelerated convergence and enhanced final model stability, as evidenced by comparative training curves. Validation through rigorous subject-wise data splits and Leave-One-Subject-Out (LOSO) cross-validation confirmed the model's ability to generalize to unseen users, a critical requirement for practical healthcare applications. Furthermore, the measured sub-50ms inference latency on embedded platforms like Raspberry Pi confirmed the framework's suitability for deployment in edge-based, real-time monitoring systems.

\subsection{Practical Applications}

The demonstrated accuracy, robustness, and low-latency performance of the 	extit{MultiModalFallDetector} framework opens several avenues for practical deployment in geriatric care and personal health monitoring. The most immediate application is within assisted living facilities and nursing homes, where the system can be integrated into wearable pendants or smartwatch-based monitoring solutions worn by residents. Upon detecting a fall with high confidence, the system can trigger an immediate alert to on-site caregivers or a centralized monitoring station, enabling rapid response that can significantly reduce the morbidity and mortality associated with prolonged lying times after a fall, often referred to as the ``long lie.''

For independent living and home monitoring scenarios, the technology can be incorporated into consumer-grade smart wearables or specialized medical alert devices. Coupled with a mobile health application, it can provide continuous, unobtrusive monitoring for elderly individuals living alone, automatically contacting emergency services or designated family members in the event of a fall. The model's ability to process multi-modal data, including physiological signals, adds a layer of contextual awareness, potentially reducing false alarms triggered by activities of daily living that mimic fall-like motion patterns, such as sitting down quickly or stumbling.

The architecture is also conducive to integration within broader smart home ecosystems and telehealth platforms. Sensor data from the wearable device could be fused with information from ambient environmental sensors (e.g., room cameras with privacy-preserving algorithms, vibration sensors on floors) within a federated learning or decision fusion framework to further improve detection accuracy and reliability. The low inference latency ensures that such integration does not compromise the system's real-time alerting capability. Furthermore, the insights gained from the auxiliary activity classifier could be leveraged to provide continuous activity profiling, offering caregivers and healthcare providers valuable longitudinal data on an individual's mobility patterns and overall health trends beyond mere fall detection.

\subsection{Future Research Directions}

Despite the promising results, this study has several limitations that delineate clear paths for future research. The most significant constraint is the reliance on laboratory-simulated fall data from the SisFall and UCI HAR datasets. While these datasets provide a controlled foundation for model development, the biomechanics and contextual factors of real-world falls, occurring on varied surfaces, involving different clothing, and accompanied by genuine surprise and fear, may differ substantially. Therefore, a paramount future direction is the collection and annotation of large-scale, real-world fall datasets from elderly populations. This requires longitudinal studies with ethical approval, involving consenting participants in their homes or care facilities, and poses significant challenges related to privacy, data labeling, and the inherent rarity of fall events.

Closely related is the issue of limited sample size and demographic diversity. The primary evaluation utilized data from 23 elderly subjects in SisFall. Future work must expand data collection to include larger, more diverse cohorts spanning a wider range of ages, genders, mobility levels (from highly active to frail), and comorbidities (e.g., Parkinson's disease, stroke history). This is essential for assessing and mitigating potential algorithmic biases to ensure the system's equitable performance across all user groups. Techniques such as domain adaptation and personalized model fine-tuning should be explored to adapt the generic model to an individual's unique movement patterns and physiological baselines.

To enhance practicality, research into online learning and adaptive personalization is warranted. A static model may not account for the gradual changes in gait and activity patterns due to aging or recovery from illness. Future systems could incorporate mechanisms for continuous, privacy-preserving learning on the edge device, allowing the detection thresholds and feature representations to adapt slowly over time to the individual user, thereby maintaining high accuracy throughout the device's lifecycle.

From a systems perspective, deeper integration with edge AI hardware presents a fruitful direction. While latency measurements on platforms like Raspberry Pi are encouraging, optimizing the model for ultra-low-power microcontrollers (MCUs) or dedicated neural processing units (NPUs) found in modern wearables would enable longer battery life and more seamless user experience. This involves exploring model compression techniques (pruning, quantization, knowledge distillation) to create a highly efficient variant of the 	extit{MultiModalFallDetector} without compromising critical detection performance.

Finally, to transition from a research prototype to a clinically validated tool, prospective clinical trials are necessary. Future studies should deploy the system in real-world settings (assisted living facilities, private homes) over extended periods to evaluate its performance, user acceptance, false alarm rate in daily life, and most importantly, its impact on health outcomes such as reduced emergency response times, fewer hospitalizations due to fall complications, and improved quality of life for elderly users and their caregivers.

\bibliographystyle{IEEEtran}
\bibliography{references}

\vspace{12pt}

\section*{Acknowledgment}

The authors would like to thank the creators of the SisFall and UCI HAR datasets for making their data publicly available, which enabled this research.

\end{document}